\documentclass[letterpaper, 10 pt, conference]{ieeeconf}
\IEEEoverridecommandlockouts
\usepackage{amsmath,amssymb,amsfonts}
\usepackage{graphicx}
\usepackage{xcolor}
\usepackage[hyphens]{url}
\usepackage[hidelinks,pdftex]{hyperref}
\usepackage[all]{hypcap}
\usepackage[T1]{fontenc}
\usepackage{multirow}

\graphicspath{{./graphics/}}

\begin{document}
\hypersetup{
	pdfinfo ={
			Title={Leveraging Procedural Generation for Learning Autonomous Peg-in-Hole Assembly in Space},
			Subject={2024 International Conference on Space Robotics (iSpaRo)},
			Author={Andrej Orsula, Matthieu Geist, Miguel Olivares-Mendez, Carol Martinez},
			Creator={LaTeX},
			Keywords={Manipulation in space; In-space manufacturing and assembly; Intelligent and autonomous space robotics systems},
		}
}

\title{\textbf{Leveraging Procedural Generation for Learning Autonomous Peg-in-Hole Assembly in Space}}


\author{
	Andrej Orsula\textsuperscript{1}
	\and
	Matthieu Geist\textsuperscript{2}
	\and
	Miguel Olivares-Mendez\textsuperscript{1}
	\and
	Carol Martinez\textsuperscript{1}
}

\maketitle\thispagestyle{empty}\pagestyle{empty}

\footnotetext[1]{
	Space Robotics Research Group~(SpaceR),
	Interdisciplinary Centre for Security, Reliability and Trust~(SnT),
	University of Luxembourg, Luxembourg.
		{\tt andrej.orsula@uni.lu}
}
\footnotetext[2]{
	Cohere, Canada.
}

\begin{abstract}
	The ability to autonomously assemble structures is crucial for the development of future space infrastructure. However, the unpredictable conditions of space pose significant challenges for robotic systems, necessitating the development of advanced learning techniques to enable autonomous assembly. In this study, we present a novel approach for learning autonomous peg-in-hole assembly in the context of space robotics. Our focus is on enhancing the generalization and adaptability of autonomous systems through deep reinforcement learning. By integrating procedural generation and domain randomization, we train agents in a highly parallelized simulation environment across a spectrum of diverse scenarios with the aim of acquiring a robust policy. The proposed approach is evaluated using three distinct reinforcement learning algorithms to investigate the trade-offs among various paradigms. We demonstrate the adaptability of our agents to novel scenarios and assembly sequences while emphasizing the potential of leveraging advanced simulation techniques for robot learning in space. Our findings set the stage for future advancements in intelligent robotic systems capable of supporting ambitious space missions and infrastructure development beyond Earth.
\textit{The source code is available at \href{https://github.com/AndrejOrsula/drl_omni_peg}{https://github.com/AndrejOrsula/drl\_omni\_peg}.}

\end{abstract}

\section{Introduction}\label{sec:introduction}

The evolving field of space robotics is a critical component for the future of space exploration, where systems capable of autonomously executing complex tasks with minimal human intervention are becoming increasingly essential. Robotic assembly and servicing stand out as key elements for the development and maintenance of infrastructure in space. These capabilities are of particular interest for planetary missions that strive to establish a sustainable human presence on other celestial bodies through initiatives like NASA's Artemis program~\cite{nasa2020artemis}. Furthermore, advancements in this domain will also shape the future of in-space assembly to facilitate the construction of large-scale orbital structures~\cite{zhihui2021review} and support in-orbit servicing activities such as refuelling~\cite{esa2023in_orbit_servicing}.

The peg-in-hole task is a fundamental manipulation skill for robots performing assembly and servicing. In space, this task is vital for the construction of modular structures as well as the repair and maintenance of satellites to extend their operational life through the replacement of faulty components or restoration of structural issues. The task is characterized by its complex contact-rich interactions that necessitate precise and robust control strategies to ensure successful completion. Traditional control methods, which often rely on detailed contact model analysis, are capable of efficiently handling well-defined scenarios but struggle to generalize across a wide range of conditions~\cite{xu2019compare}. This limitation can be particularly apparent in space due to the unpredictable and highly variable conditions that might include part and material inconsistencies caused by post-launch damage, thermal expansion and long-term exposure to the harsh conditions of space. Therefore, the capacity to generalize and adapt to novel scenarios becomes crucial for autonomous robotic systems operating in these environments.

Reinforcement learning (RL) offers a promising approach to these challenges by providing a formal framework to optimize policies for sequential decision-making problems. In particular, deep RL has shown remarkable success in learning diverse tasks from high-dimensional sensory inputs and adapting to previously unseen scenarios. These qualities have also made RL attractive for tackling complex contact-rich manipulation tasks, including peg-in-hole assembly~\cite{elguea2023review}. However, the deployment of RL in real-world robotic applications remains challenging, especially in space where the limited accessibility to physical systems restricts the training and validation of RL agents on the actual hardware. As a consequence, the reliance on realistic simulation environments has become increasingly evident, with the aim of narrowing the sim-to-real gap. Nonetheless, the generalization of RL agents to novel scenarios is still a topic of active research.

\begin{figure}[t]
    \vspace{2.057mm}
    \centering
    \includegraphics[width=1.0\linewidth]{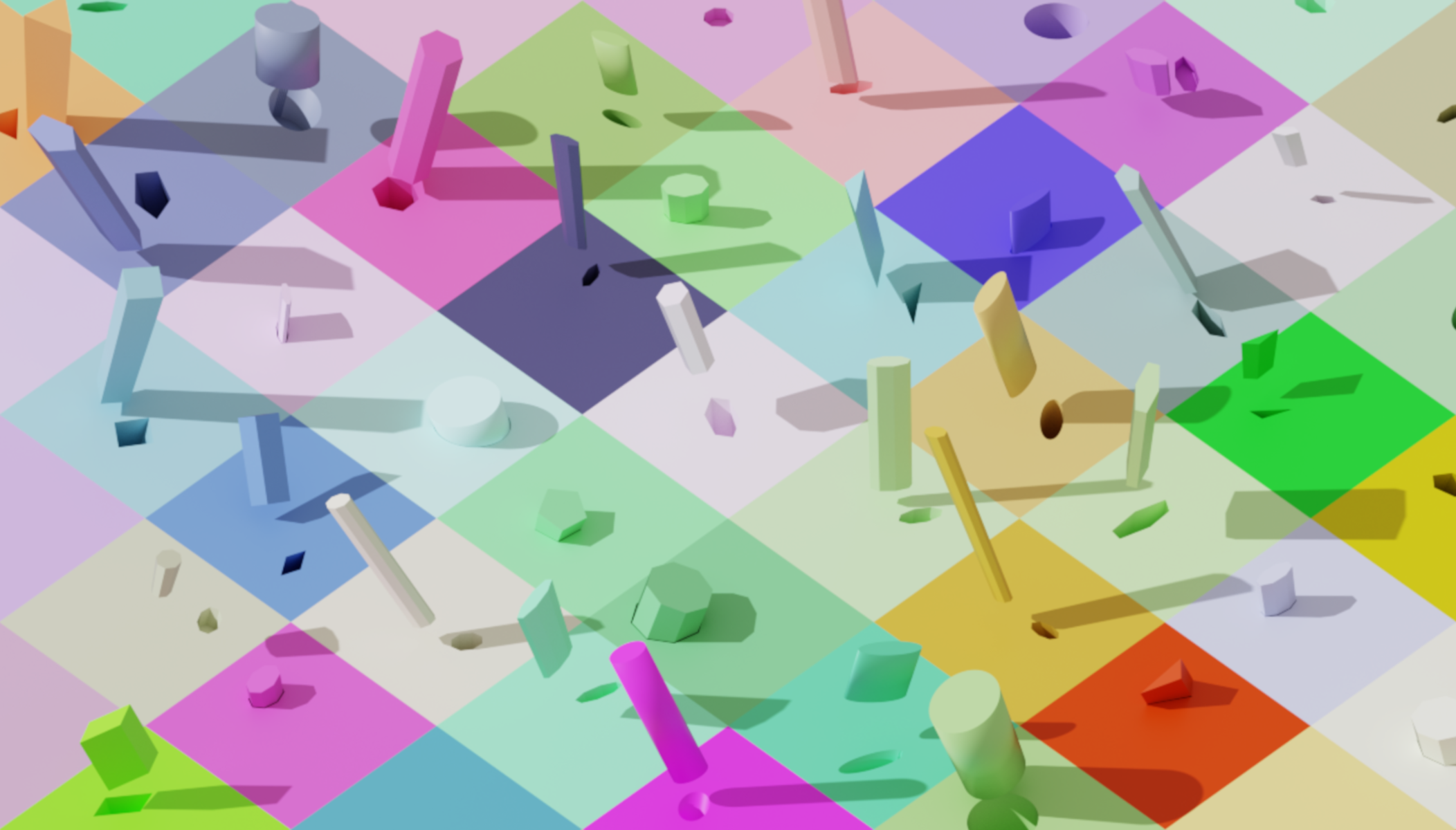}
    \caption{Our approach leverages procedural generation and domain randomization in a highly parallelized simulation environment to train agents that can generalize across a wide range of peg-in-hole assembly scenarios.}
    \label{fig:front-page}
    \vspace{-1.0\baselineskip}
\end{figure}

This work introduces an approach for learning autonomous peg-in-hole assembly with an emphasis on improving the ability to generalize across different scenarios. Our contribution is based on leveraging procedural generation to create a diverse set of peg-in-hole modules, each providing a unique scenario for comprehensive training of the agents. These modules are incorporated into a highly parallelized simulation environment, enabling agents to efficiently collect a wide range of experience across a broad spectrum of assembly scenarios. Furthermore, domain randomization techniques are applied within the simulation environment to enhance the robustness and generalization of the agents. With this framework, we train and evaluate agents using three distinct RL algorithms to explore the benefits of different learning paradigms on contact-rich manipulation tasks:~model-free on-policy PPO, model-free off-policy SAC, and model-based DreamerV3. Additionally, we investigate the impact of temporal dependencies on the successful learning of the peg-in-hole assembly task. The study concludes by demonstrating the adaptability of the trained agents to a novel assembly sequence, highlighting the potential of our approach.

\section{Related Work}\label{sec:related-work}

Autonomous robotic assembly, characterized by its requirements for precision, robustness and adaptability, poses considerable challenges across both terrestrial and extraterrestrial applications. The peg-in-hole task, in particular, has been a focal point for research in this domain and has been explored through various control strategies and learning-based approaches.

Historically, the field has leaned towards classical strategies based on identifying contact states through model recognition and subsequent use of compliant control with pre-programmed trajectories, such as spiral search, to guide the peg into the hole~\cite{jasim2017contact,lee2022contact}. However, these strategies usually assume near-perfect alignment between the peg and its hole before the insertion can begin, requiring either extensive pre-calibration or additional alignment procedures~\cite{tang2016autonomous}. Furthermore, many existing methods cannot handle certain conditions, notably holes that are not perpendicular to their mounting surfaces. Although these methods have shown success in well-defined scenarios and provide explicit safety guarantees, they often struggle to adapt to novel scenarios~\cite{xu2019compare}, such as the unpredictable conditions of space environments. This limitation has motivated the exploration of learning-based approaches to address these challenges.

As a step towards more versatile approaches, learning from demonstration has been successfully applied to several robotic assembly tasks, including the peg-in-hole assembly. This approach enables robots to learn complex manipulation skills by observing and interpreting human demonstrations. Techniques leveraging Gaussian-based regression models~\cite{tang2016teach,wan2017optimal} have shown promise in mimicking the compliant behaviours of human experts. However, the effectiveness of these methods is bounded by the quantity and diversity of the provided demonstrations, constraining their ability to generalize to novel scenarios not present in the training data. This limitation motivated the exploration of methods combining imitation learning to acquire an initial policy and RL for further refinement~\cite{cho2020learning,wang2023learning}. Although these methods have demonstrated robust behaviour without explicit programming, their generalization capabilities might still be restricted to a small set of peg-in-hole scenarios due to potential biases present in the demonstration data.

Recent advancements in RL have marked a significant shift in the paradigm for learning contact-rich manipulation tasks. Most research efforts have focused on applying deep RL to robotic assembly, while other tasks, such as door opening and folding of deformable fabrics, have also been explored~\cite{elguea2023review}. The peg-in-hole assembly task is among the most commonly researched applications of RL in contact-rich manipulation, with various innovative methods being proposed to address the challenges of this task. The use of model-free RL algorithms is predominant in the literature~\cite{inoue2017deep,beltran2020variable,li2022multiple,wang2022deep}, driven by the challenges of accurately modelling the complex interaction dynamics. Precise modelling of interactions between the peg and hole can be crucial, as even a slight misalignment of the peg can lead to significant friction and impact the success of the task. Although less common, the adoption of model-based RL strategies has demonstrated the potential to enhance sample efficiency through the use of learned dynamics models~\cite{luo2018deep,thomas2018learning,ding2019transferable}. Recurrent policies have also been explored to handle the temporal dependencies of assembly tasks~\cite{inoue2017deep,wang2022deep}, especially those utilizing Long Short-Term Memory networks~\cite{hochreiter1997long}.

Sensory input, included as part of the observation space, plays a crucial role in the learning process. Wrist-mounted force-torque sensors have been widely adopted due to their ability to provide important feedback during contact-rich manipulation tasks such as peg-in-hole assembly~\cite{inoue2017deep,beltran2020variable,luo2018deep,thomas2018learning,ding2019transferable}. They enable robots to adjust their actions in real-time to successfully guide the peg into the hole, although both parts must already be in contact for useful data to be obtained. Vision-based perception has also been explored to provide estimates of the relative pose between the peg and the hole~\cite{wang2022deep}; however, most approaches employ visual perception only for initial rough alignment due to the challenges posed by the complex interactions~\cite{thomas2018learning,ding2019transferable}. Similarly, the selection of control strategies has been diverse but has predominantly focused on hybrid position-force control~\cite{inoue2017deep,beltran2020variable} and admittance control~\cite{luo2018deep}. As many robotic manipulators are not inherently compliant and offer only position and velocity control interfaces, position control in the form of end-effector displacement has also been explored~\cite{wang2022deep,ding2019transferable}. Our approach takes a different perspective by focusing on velocity control in the Cartesian space without relying on force-torque sensors. This facilitates generalization across a wide range of actuation mechanisms while alleviating the need for specific sensor technologies.

Despite notable progress, the generalization of RL agents to novel peg-in-hole scenarios remains an open challenge. Current research predominantly focuses on a small number of pegs that are mostly cylindrical and are often restricted to a tabletop setting with constrained degrees of freedom. Some studies have revolved around specific cases, exemplified by the simultaneous insertion of multiple pegs~\cite{li2022multiple} and deformable holes~\cite {luo2018deep}. However, despite efforts to evaluate generalization capabilities, such as the insertion of USB and LAN connectors~\cite{beltran2020variable}, the research in this area is limited and has encountered reduced success rates. This situation signifies the need for approaches that can improve the generalization capabilities of RL agents. In pursuit of this goal, our work seeks to contribute to the field by exploring the potential of improving the generalization capabilities of RL agents performing peg-in-hole assembly through the use of procedural generation and domain randomization techniques~\cite{tobin2017domain}. By diversifying the training experience, we aim to take the first steps towards RL agents that can succeed in a broader spectrum of peg-in-hole assembly scenarios and contribute to the development of more adaptable autonomous robotic systems.

\section{Problem Statement}\label{sec:problem-statement}

The task of peg-in-hole assembly embodies a fundamental challenge in autonomous robotics, especially within the demanding and unpredictable conditions of space environments. In this domain, it is essential for operations such as modular assembly of orbital structures and maintenance of infrastructure. Therefore, the development of autonomous systems capable of performing peg-in-hole assembly tasks with a high degree of generalization and robustness is crucial for the success of future missions. The peg-in-hole task generally involves the insertion of a peg into a corresponding hole that is slightly larger than the peg itself, with an accommodating clearance margin. The complexity of this process can be affected by several factors, including the peg's shape and size, the hole's orientation, and the material properties of the mating surfaces, all while requiring precise alignment and insertion.

In this investigation, we predominantly consider pegs with arbitrary convex geometry to cover a wide range of realistic scenarios. The bottom and top of the pegs are assumed to be flat and normal to the peg's axis. Each peg has exactly one corresponding hole, which is also assumed to have a flat bottom that is normal to the axis of the hole. The orientation of the hole is not necessarily perpendicular to its mounting surface, which increases the complexity of the task by forcing the peg to be inserted at varying angles for successful assembly. The peg and hole are assumed to be rigid, and the task is defined with a specific focus on a radial clearance of~1.0~mm between the peg and the hole. The magnitude of this clearance is selected based on the envisioned use-case of in-orbit modular assembly, where such margin would be necessary to accommodate variances from environmental factors like thermal expansion and potential manufacturing processes, including in-orbit additive manufacturing. This consideration is vital for ensuring the feasibility and reliability of autonomous assembly operations in the challenging environment of space. Lastly, the edges of pegs and holes are assumed to be sharp, without any chamfers, fillets, or other mechanical design features that would otherwise ease the insertion process.

To limit the complexity of the task and focus on the core challenges of the peg-in-hole assembly without the influence of additional factors, we constrain our analysis to simulations and avoid detailed control and perception mechanisms. Instead, we focus on direct control of the peg's trajectory within Cartesian space. This methodological choice facilitates adaptability to different platforms across a broad spectrum of actuation mechanisms and the application of established control techniques capable of following a trajectory. Similarly, we only consider the explicit knowledge of the relative pose between the peg and the hole without the use of any additional sensory information about contact forces or dimensions of the peg. Upon deployment, the relative pose could be estimated through the use of vision-based perception, but the approach would not necessitate the use of other specific sensor technologies. This decision is motivated not only by the desire to reduce the potential sources of uncertainties but also to ensure the general applicability of the approach across a wide range of platforms. Nevertheless, to simulate the inherent uncertainties in the perception and control systems encountered in real-world scenarios, we introduce noise to both the observed state and the executed control commands. This noise is modelled to reflect a normal distribution, with a standard deviation set at~0.5~mm for position and~1.0\textdegree\ for orientation, reflecting the precision levels achievable with modern vision-based systems designed for proximity operations.

\section{Learning Autonomous Peg-in-Hole Assembly}\label{sec:learning-autonomous-peg-in-hole-assembly}

To address the challenges of autonomous peg-in-hole assembly, we adopt a learning-based approach that leverages diverse training scenarios to enhance the generalization capabilities of RL agents. Central to our methodology is the procedural generation of assembly modules, which is combined with a highly parallelized simulation environment and domain randomization techniques to provide a rich source of experience for the agents. This section outlines its key components, including the task formulation, procedural generation of assembly modules, and the simulation environment.

\subsection{Task Formulation}\label{ssec:task-formulation}

Our approach to autonomous peg-in-hole assembly can be formulated as a partially observable Markov decision process (POMDP) that encapsulates the sequential decision-making nature of the task. Partial observability arises from the agent's limited knowledge of the current state, particularly about the contact forces as well as the shape, dimensions, and material properties of the peg and hole. The objective of the agent is to learn an optimal policy that maximizes the expected cumulative reward over time, thereby achieving a successful peg insertion. Formulating the task is thus critical for applying RL, as it clearly defines the interactions between the agent and its environment through observations, actions, and a reward function.

\subsubsection*{Observation Space}\label{sssec:observation-space}

\begin{figure}[ht]
    \centering
    \includegraphics[width=1.0\linewidth]{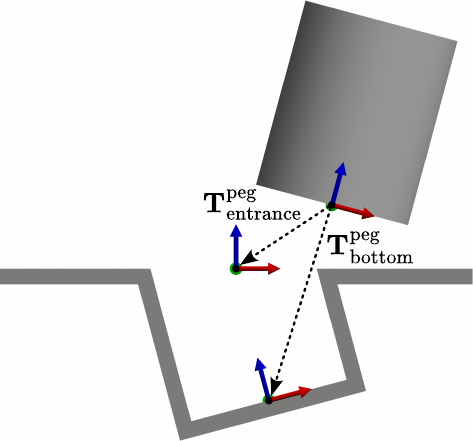}
    \caption{The observation space of our agents is defined by the~\(\mathbf{T}_{\text{entrance}}^{\text{peg}}\) and~\(\mathbf{T}_{\text{bottom}}^{\text{peg}}\) transformations that capture the spatial state of the assembly.}
    \label{fig:observation-space}
\end{figure}

The observations provided to the agent capture the spatial state of the peg-in-hole assembly. Specifically, the observation space is structured around two key transformations visualized in Fig.~\ref{fig:observation-space}; one from the bottom of the peg to the hole entrance~\(\mathbf{T}_{\text{entrance}}^{\text{peg}}\) and another from the bottom of the peg to the bottom of the hole~\(\mathbf{T}_{\text{bottom}}^{\text{peg}}\). The frame at the hole entrance is oriented with respect to the normal vector of the hole's mounting surface, while the frame at the bottom of the hole is oriented based on the normal vector of the hole's bottom surface. The translation for each transformation is represented as relative \(x\),~\(y\),~and~\(z\) coordinates, whereas the orientation is encoded using a~6D~representation~\cite{zhou2019continuity} derived from the first two columns of the rotation matrix. Each transformation contributes nine values, with three values for translation and six for orientation, resulting in a combined 18-dimensional observation vector.

\subsubsection*{Action Space}\label{sssec:action-space}

The actions that the agent can select to interact with the environment are designed to provide direct control over the peg's movement, enabling precise manoeuvres required for successful assembly. The action space comprises a continuous six-dimensional vector, split evenly between the target linear and angular velocities of the peg in Cartesian space. The linear velocities are aligned with the \(x\),~\(y\),~and~\(z\) axes, while the angular velocities control the peg's rotation about these axes, with the central point of rotation at the bottom of the peg. All actions are clipped to the range of~\([-1.0, 1.0]\) and mapped to their real-world equivalents, with linear velocities scaled to a maximum of~25.0~cm/s and angular velocities to a maximum of~90.0\textdegree/s. This scheme allows for a comprehensive range of movement and orientation adjustments, giving the agent the flexibility to develop sophisticated strategies for peg insertion.

\subsubsection*{Reward Function}\label{sssec:reward-function}

We incentivize the agent's learning process through a dense reward function that provides continuous feedback on the agent's actions. It integrates the positional and angular distance based on the transformation from the bottom of the peg to the bottom of the hole~\(\mathbf{T}_{\text{bottom}}^{\text{peg}}\). The positional component of the reward is determined by the Euclidean distance between these two frames, and the orientation component is calculated as the norm of the difference between the current and the target rotation matrix of the peg that would align the peg correctly with the hole. The reward function assigns a weight of~0.8 to the positional component and~0.2 to the orientation component to emphasize the prioritization of bringing the peg close to the hole before fine-tuning its orientation for insertion.

The computation of the reward is relative to the initial pose of the peg at the start of each episode, where a positive reward is granted for actions that reduce the distance up to a maximum cumulative reward of~1.0. Similarly, a negative reward is returned for actions that increase the distance. Furthermore, a fixed negative reward of~\(-\)1.0 is granted in cases where the peg moves below the hole's mounting surface in order to prevent the agent from attempting to insert the peg from the incorrect side. The episode is terminated in both events when the peg is successfully inserted into the hole or when the peg moves below the hole's mounting surface. This reward function plays an important role in guiding the RL agent's learning process towards successful peg insertion.

\subsection{Procedural Generation of Assembly Modules}\label{ssec:procedural-generation-of-assembly-modules}

The procedural generation of assembly modules is a key component of our approach in providing a diverse set of training scenarios that expose the RL agents to a wide range of different experiences. This diversity is crucial for enhancing the generalization capabilities of the agents, preparing them for the unpredictable conditions of space environments. Analogous to our previous approach for generating procedural lunar terrains and rocks~\cite{orsula2022learning}, we employ Blender~\cite{blender} with its Geometry Nodes feature to create pipelines that enable the dynamic creation of a nearly infinite variety of peg-in-hole modules, each with unique configurations of pegs and corresponding holes.

The pipelines are based on the systematic creation and transformation of mesh geometry through parametric operations, allowing for the generation of a wide range of peg and hole geometries. The initial shape of each peg is determined by its horizontal cross-section, which can range from a simple polygon to a circle, depending on the number of specified vertices. This cross-section is configurable both in its circumradius and aspect ratio along any axis while maintaining convexity across all variations. The 2D cross-sections are extruded to form 3D pegs with a configurable height, where some pegs can be tapered down by downscaling the cross-section towards their bottom, further enhancing the diversity of the pegs. The holes are derived from their corresponding peg geometries on top of a~15\texttimes15~cm flat square module with a configurable depth of each hole. The reconfigurability also extends to the 2D position and 3D orientation of the holes, with the latter necessitating the peg to be inserted at a potentially non-perpendicular angle relative to the module's surface. Likewise, the clearance between the peg and the hole is adjustable, which we fixate to~1.0~mm as per Section~\ref{sec:problem-statement}.

\begin{figure}[ht]
    \vspace{-1.0\baselineskip}
    \centering
    \includegraphics[width=1.0\linewidth]{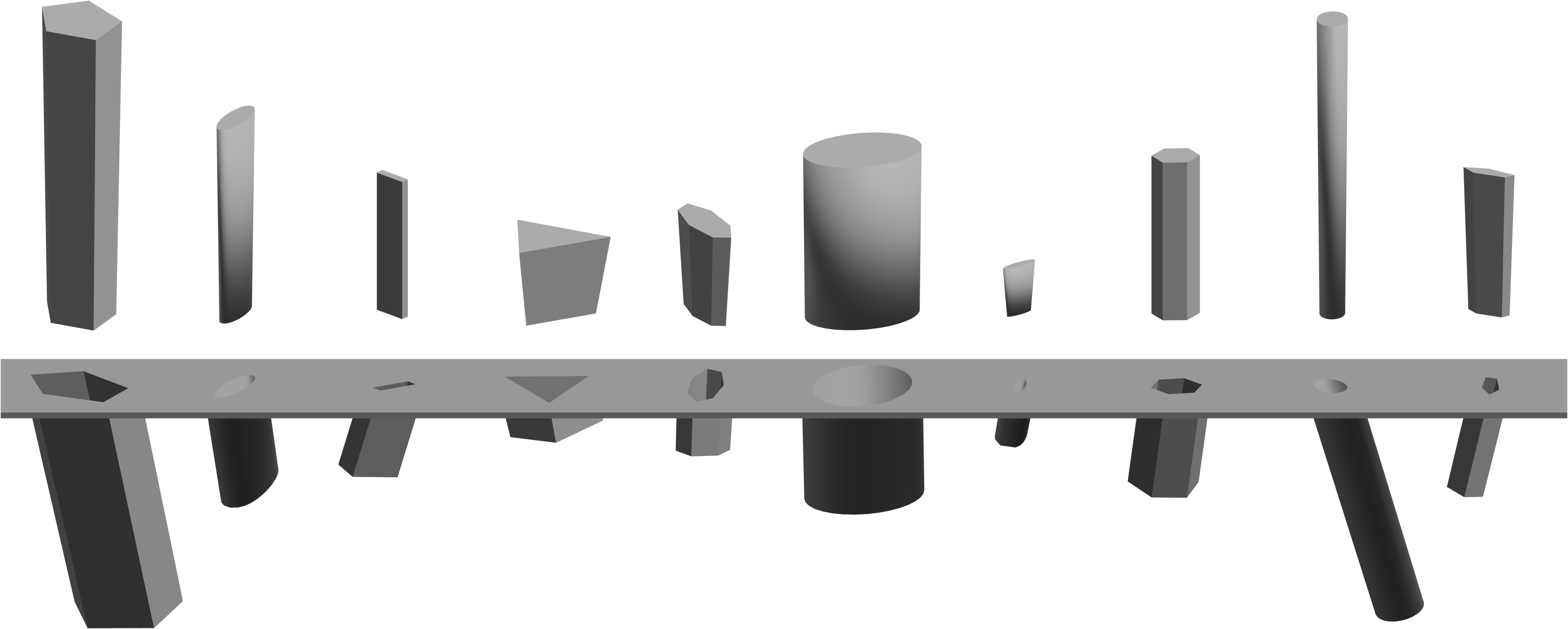}
    \caption{Ten samples of the procedurally generated peg-in-hole modules, illustrating the diversity of the peg and hole geometries.}
    \label{fig:procedural-generation}
\end{figure}

Rather than manually specifying all the parameters for each generated peg-in-hole module pair, we randomize the generation process to create a broad spectrum of peg-in-hole modules. Each parameter is sampled from a uniform distribution within a specified range, ensuring that the generated modules exhibit a variety of characteristics within the defined constraints. The configuration selected during our experimental evaluation is detailed in Table~\ref{tab:procedural-generation}, with samples of the corresponding procedurally generated peg-in-hole modules shown in Fig.~\ref{fig:procedural-generation}. The resulting pegs and holes, with their randomized characteristics, contribute to a robust training environment that fosters the generalization capabilities of our agents. Moreover, the design of these modules facilitates their direct translation into physical models through additive manufacturing, exemplified by 3D printing.

\begin{table}[ht]
    \centering
    \caption{The Parameter Configuration for the Procedural Generation of Diverse Peg-in-Hole Modules}
    \label{tab:procedural-generation}
    \begin{tabular}{lr@{\hspace{3pt}}l}
        \hline
        \multicolumn{1}{c}{\textbf{Parameter}}      & \multicolumn{2}{c}{\textbf{Range}}\hspace{0.25pt}                                 \\
        \hline
        \multirow{2}{*}{Peg cross-section vertices} & \(75\)\% \(3\)                                    & -- \(8\)                      \\
                                                    & \(25\)\% \hspace{0.5em}                           & \hspace{-5pt} \(32\)          \\
        Peg cross-section circumradius              & \(1.0\)                                           & -- \(3.0\) cm                 \\
        Peg cross-section aspect ratio              & \(0.25\)                                          & -- \(1.0\)                    \\
        Peg height                                  & \(2.5\)                                           & -- \(15.0\) cm                \\
        Peg tapering                                & \(0.0\)                                           & -- \(25.0\)\%                 \\
        Hole depth (fraction of peg height)         & \(40.0\)                                          & -- \(80.0\)\%                 \\
        Hole orientation                            & \(-15.0\)                                         & -- \(15.0\text{\textdegree}\) \\
        \hline
    \end{tabular}
\end{table}

\subsection{Simulation Environment}\label{ssec:simulation-environment}

Our simulation environment is built on top of NVIDIA Omniverse~\cite{omniverse} to provide a scalable and realistic platform for training and evaluating RL agents. This selection is motivated by its high-fidelity physics engine that supports Signed Distance Field collision checking~\cite{macklin2020local}, which provides stable and accurate simulations of contact-rich interactions found in the peg-in-hole assembly tasks. Furthermore, the capability to simulate a vast array of assembly scenarios in parallel is instrumental for the agents to efficiently collect a wide range of experiences to enhance their generalization capabilities. The environment is shown in Fig.~\ref{fig:front-page}, with several peg-in-hole modules being simulated in parallel.

The simulation environment is populated with a configurable number of unique procedurally generated peg-in-hole modules, where each module is treated as an independent environment worker. The agents interact with these modules through the observations and actions defined in Section~\mbox{\ref{ssec:task-formulation}}. At the same time, the environment provides the corresponding rewards and terminates the episodes based on the defined criteria to indicate success or failure. Our agents operate with a common control frequency of~50~Hz while the physics scene is updated at~200~Hz to maintain a high level of fidelity in the simulation. The motion of each simulated peg is maintained through a PD controller that tracks the target velocities provided by the agents, ensuring that the pegs move smoothly and responsively. The modules with holes are static, and the effect of gravity on the pegs is disabled, which mimics the conditions of a secure grasp by a robotic manipulator mounted on the same rigid body as the module.

We model the peg-in-hole assembly task as an episodic problem, with each episode terminated after a maximum of~500 timesteps, equalling~10~s. This duration is selected to ensure that the agents have sufficient time to complete the assembly task while maintaining a balance between the training efficiency and the computational cost of the simulations. At the beginning of each episode, the pegs are spawned at uniformly sampled random positions and orientations within an area of~50\texttimes50~cm above their corresponding holes. Similarly, the physical properties of the materials used in each peg-in-hole module pair, such as friction and restitution coefficients, are also randomized at the start of each episode. This variability further challenges the agents via the process of domain randomization, encouraging them to adapt their strategies to different scenarios that mirror the diversity of conditions they would encounter in the actual task.

\section{Experimental Results}\label{sec:experimental-results}

The experimental evaluation focuses on assessing the effectiveness of our approach for autonomous peg-in-hole assembly, with a particular emphasis on the generalization capabilities of RL agents across a spectrum of novel scenarios. This section outlines the evaluation methodology and the key findings from our experiments.

\subsection{Evaluated Reinforcement Learning Algorithms}\label{ssec:evaluated-reinforcement-learning-algorithms}

We extend our evaluation to three distinct RL algorithms, each representing a different approach within the spectrum of RL methodologies.

\textit{Proximal Policy Optimization (PPO)}~\cite{schulman2017proximal} is a model-free on-policy algorithm based on policy gradients optimized through a clipped surrogate objective to ensure stable policy updates. It has gained widespread adoption due to its robust performance across various domains, including continuous control in robotics.

\textit{Soft Actor-Critic (SAC)}~\cite{haarnoja2018soft} is a model-free off-policy algorithm that employs an actor-critic framework with entropy regularization to encourage exploration. It has gained significant attention due to its performance in continuous action spaces.

\textit{DreamerV3}~\cite{hafner2023mastering} is a model-based algorithm that concurrently learns a world model and employs this model's latent representation with a recurrent state to generate abstract sequences that are then used to optimize actor and critic networks. It has demonstrated a capacity for solving various tasks in diverse domains, including robotic manipulation.

The rationale behind evaluating agents trained using these three algorithms is to explore the spectrum of RL methodologies in the context of the peg-in-hole assembly task. By comparing these approaches, we aim to uncover insights into their respective strengths and weaknesses while also gaining a better understanding of the task itself.

We also investigate the effect of temporal dependencies on the learning process due to the POMDP nature of our task formulation. To this end, we introduce two additional variants of agents, namely PPO-STACK and SAC-STACK, incorporating observation stacking~\cite{mnih2015human} with a length of~10 to provide the agents with a history of observations that can enable them to infer unobserved states, such as the geometry and material properties of the peg and hole. DreamerV3 inherently incorporates a recurrent state in its world model and thus does not require an additional variant to capture temporal dependencies.

\subsection{Training Process}\label{ssec:training-process}

We employ our simulation environment with~1024 parallel workers and an equal number of unique peg-in-hole modules to train all five variants of the RL agents. These workers are synchronized to interact with the modules and collect experience in parallel, while a single learner processes updates to the policy networks based on the collected experience. The list of relevant non-default hyperparameters for each algorithm used during the training is detailed in Table~\ref{tab:hyperparameters}. All agents are trained for a total of~100 million steps, which corresponds to approximately~23 days of simulated experience. We integrate a simple curriculum strategy that spans the first~50 million steps and linearly increases the uniformly sampled range of initial positional and angular distance of the peg from the hole entrance to facilitate the learning process through simplified exploration, after which full randomization of initial conditions is employed.

\begin{table}[ht]
    \centering
    \caption{Selected Hyperparameters for the Training of Agents}
    \label{tab:hyperparameters}
    \resizebox{\columnwidth}{!}{%
        \begin{tabular}{lccc}
            \hline
            \multicolumn{1}{c}{\textbf{Parameter}} & \textbf{PPO}                                       & \textbf{SAC}                                       & \textbf{DreamerV3}                                                                             \\
            \hline
            Architecture (all networks)            & \([512, 512]\)                                     & \([512, 512]\)                                     & \([512, 512]\)                                                                                 \\
            Learning rate                          & \(3 \hspace{-0.2em}\cdot\hspace{-0.25em} 10^{-4}\) & \(8 \hspace{-0.2em}\cdot\hspace{-0.25em} 10^{-5}\) & \begin{tabular}[c]{@{}c@{}}World: \(10^{-4}\)\\ Actor-Critic: \(3 \cdot 10^{-5}\)\end{tabular} \\
            Discount factor                        & \(0.997\)                                          & \(0.997\)                                          & \(0.997\)                                                                                      \\
            Batch size                             & \(8192\)                                           & \(4096\)                                           & \(16\), Length: \(64\)                                                                         \\
            Horizon                                & \(128\)                                            & N/A                                                & Imagination: \(25\)                                                                            \\
            Entropy coefficient                    & \(3 \hspace{-0.2em}\cdot\hspace{-0.25em} 10^{-4}\) & \(3 \hspace{-0.2em}\cdot\hspace{-0.25em} 10^{-4}\) & \(3 \hspace{-0.2em}\cdot\hspace{-0.25em} 10^{-4}\)                                             \\
            Num. epochs / train ratio              & \(8\)                                              & \(8\)                                              & \(8\)                                                                                          \\
            Replay buffer size                     & N/A                                                & \(2 \hspace{-0.2em}\cdot\hspace{-0.25em} 10^7\)    & \(2 \hspace{-0.2em}\cdot\hspace{-0.25em} 10^7\)                                                \\
            Replay buffer warm-up                  & N/A                                                & \(5 \hspace{-0.2em}\cdot\hspace{-0.25em} 10^5\)    & \(5 \hspace{-0.2em}\cdot\hspace{-0.25em} 10^5\)                                                \\
            \hline
        \end{tabular}
    }
\end{table}

\begin{figure}[ht]
    \centering
    \includegraphics[width=1.0\linewidth]{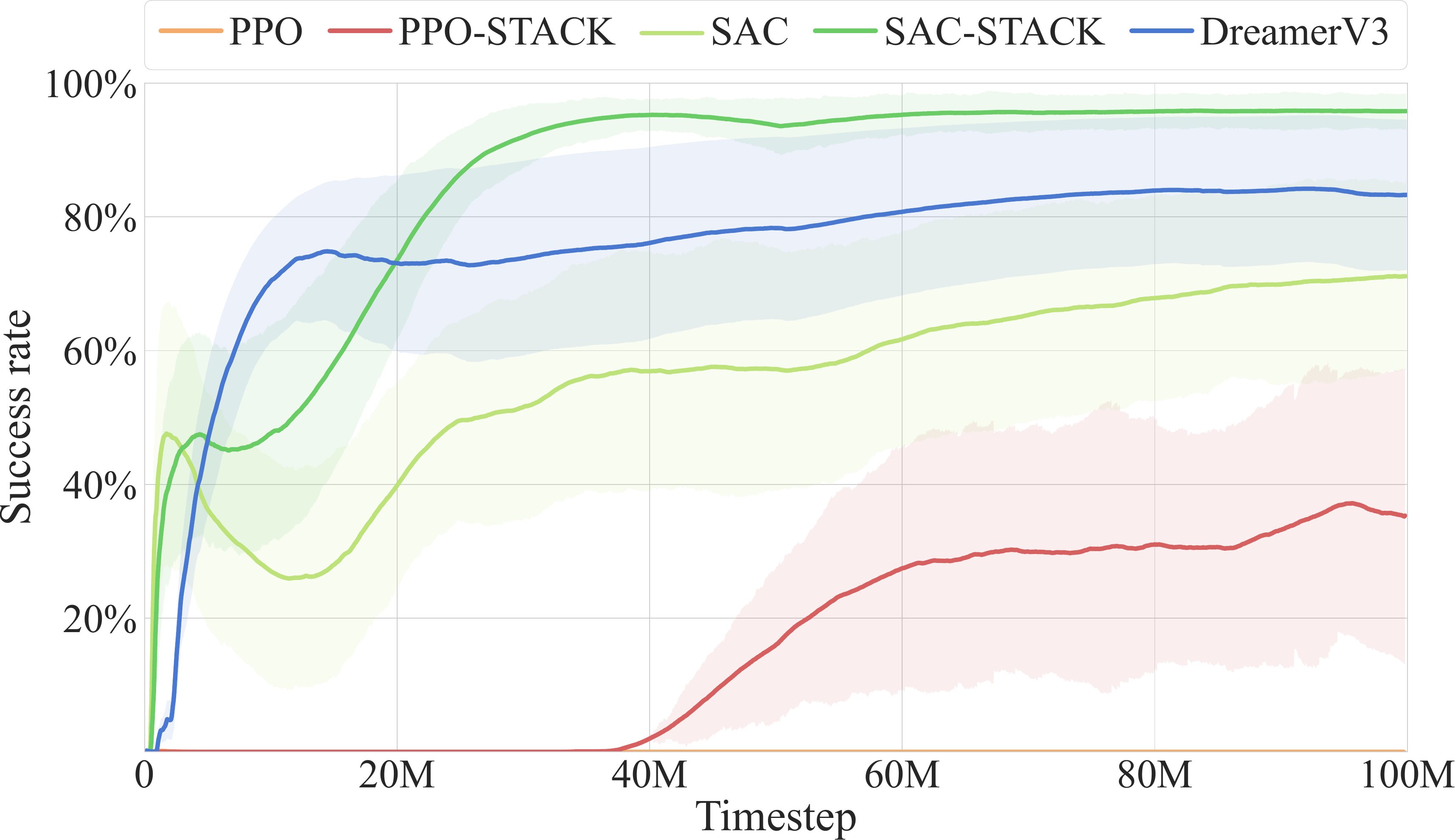}
    \caption{Learning curves of agents during the training, depicting their mean success rate with the shaded areas representing the standard deviation across three random seeds. The data is sourced directly from the training rollouts of agents, in which they follow their exploratory policies.}
    \label{fig:learning-curves}
\end{figure}

All variants of agents are trained from scratch with three distinct pseudo-random seeds to ensure the statistical significance of the results. The learning curves of the agents during the training process are presented in Fig.~\ref{fig:learning-curves}, illustrating the mean success rate of the agents over the course of their training. It is evident that both SAC and DreamerV3 exhibit a significantly faster convergence rate compared to PPO. Neither of the three PPO runs was able to achieve a successful peg insertion, while only one of the three PPO-STACK runs was able to explore the reward associated with successful peg insertion and learn to exploit it at a steady rate. Instead, all other PPO and PPO-STACK runs eventually converged to a local minimum where they would approach and hover over the hole entrance without attempting to insert the peg. In contrast, both variants of SAC were able to achieve a considerable success rate, with SAC-STACK showing a more stable learning curve and a considerably higher success rate compared to SAC. DreamerV3 also demonstrated its sample efficiency by achieving a high success rate within the first ten million steps and then steadily improving throughout the rest of the training process.

\subsection{Evaluation on Novel Peg-in-Hole Modules}\label{ssec:evaluation-on-novel-peg-in-hole-modules}

To evaluate the generalization capabilities of the trained agents, we assess their performance on two distinct sets of peg-in-hole modules. The first set consists of~1024 modules that were used during the training phase, while the second set consists of another~1024 peg-in-hole modules that are distinct from those encountered during the training phase. For each agent variant, we consider the final policy networks of all three seeds and evaluate their performance over the course of~10 attempts per module, resulting in a total of~30720 evaluated episodes for each agent variant on each set of peg-in-hole modules. Similar to the training phase, noise is applied to both the observations and actions of the agents to simulate the inherent uncertainties in perception and control systems. Furthermore, we also evaluate a random agent with a uniform distribution of actions to serve as an indicator of the inherent exploration challenges of the task.

\begin{table}[ht]
    \centering
    \caption{Success Rate of Agents During the Evaluation}
    \label{tab:success-rate}
    \begin{tabular}{lcc}
        \hline
        \textbf{Agent} & \textbf{Training Set}                       & \textbf{Test Set}                           \\
        \hline
        Random         & \hspace{4pt}\(0.00\)\%                      & \hspace{4pt}\(0.00\)\%                      \\
        PPO            & \hspace{4pt}\(0.00\)\%                      & \hspace{4pt}\(0.00\)\%                      \\
        PPO-STACK      & \(20.52\)\%                                 & \(20.10\)\%                                 \\
        SAC            & \(68.53\)\%                                 & \(49.73\)\%                                 \\
        SAC-STACK      & \(82.97\)\%                                 & \(54.70\)\%                                 \\
        DreamerV3      & \hspace{1.5pt}\(\mathbf{94.32}\)\textbf{\%} & \hspace{1.5pt}\(\mathbf{93.94}\)\textbf{\%} \\
        \hline
    \end{tabular}
\end{table}

The success rates are summarized in Table~\ref{tab:success-rate}. With our POMDP formulation and the inherent challenges of the peg-in-hole task, PPO agents were unable to achieve successful peg insertion in either of the evaluated sets. The PPO-STACK variant, which incorporates observation stacking, was able to achieve a modest success rate of~20.52\% on the training set, with a slightly worse performance on the test set. Both SAC and SAC-STACK agents demonstrated a high success rate of~68.53\% and~82.97\% on the training set, respectively. However, their performance dropped significantly on the test set, with the SAC-STACK variant achieving a higher success rate of~54.70\%. DreamerV3 agents achieved the highest success rate of~94.32\% on the training set, and their performance remained high at~93.94\% on the test set.

\begin{figure}[ht]
    \centering
    \includegraphics[width=1.0\linewidth]{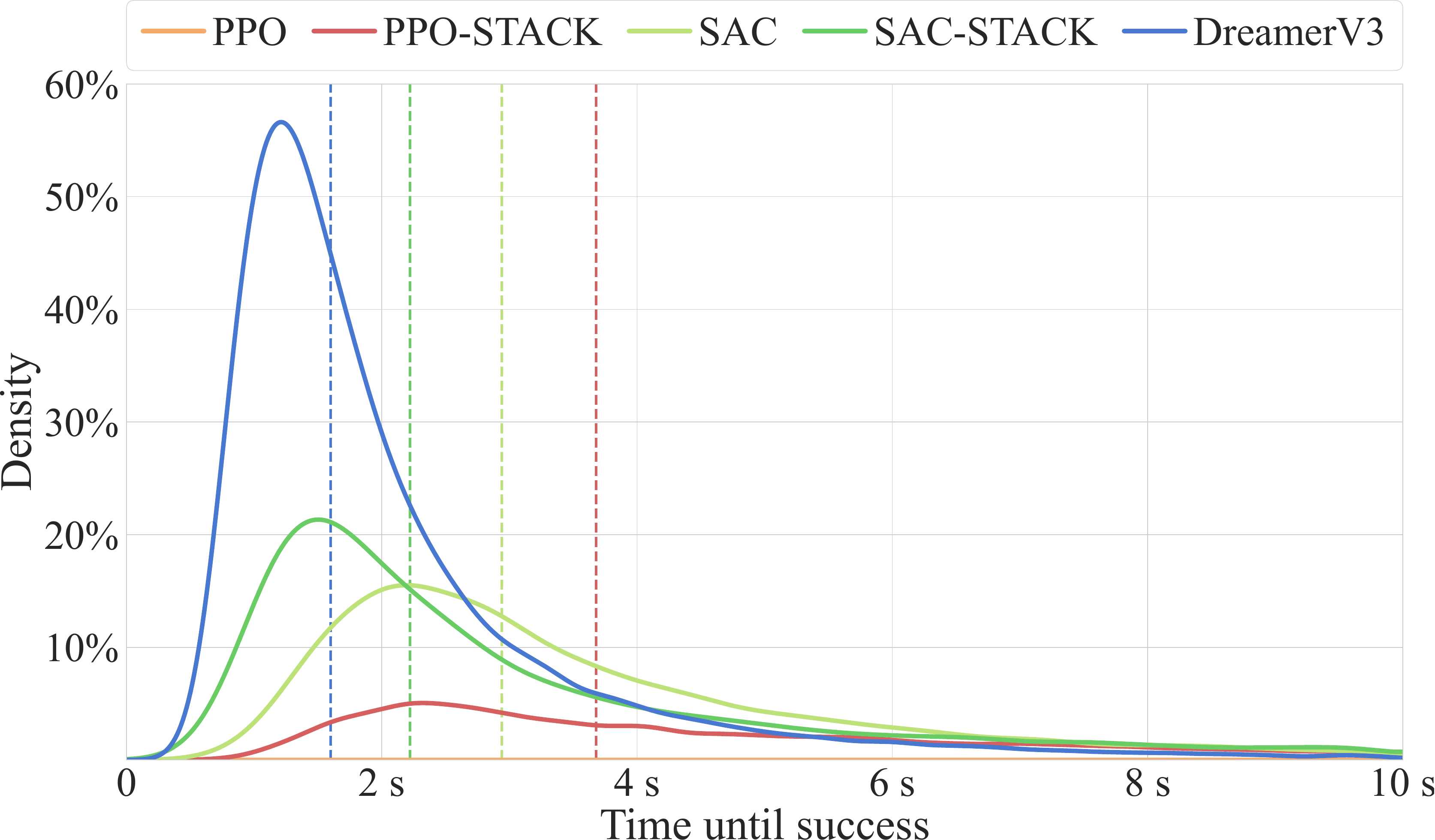}
    \caption{Distribution of the time until successful completion of novel peg-in-hole modules. The dashed lines represent the median time until completion.}
    \label{fig:time-until-completion}
\end{figure}

In addition to the success rates, we also analyze the response time of the agents to provide further insights into their efficiency. The distribution of the time until their successful completion on the test set is visualized in Fig.~\ref{fig:time-until-completion}. DreamerV3 exhibits the fastest response with a median time until successful completion of only~1.60~s, followed by SAC-STACK with a median time of~2.22~s. This result signifies that agents are capable of efficiently completing the task without requiring any additional objectives or constraints.

\subsection{Assembly Task Sequence}\label{ssec:assembly-task-sequence}

\begin{figure}[ht]
    \centering
    \begin{minipage}[b]{0.45\linewidth}
        \centering
        \includegraphics[width=1.0\linewidth]{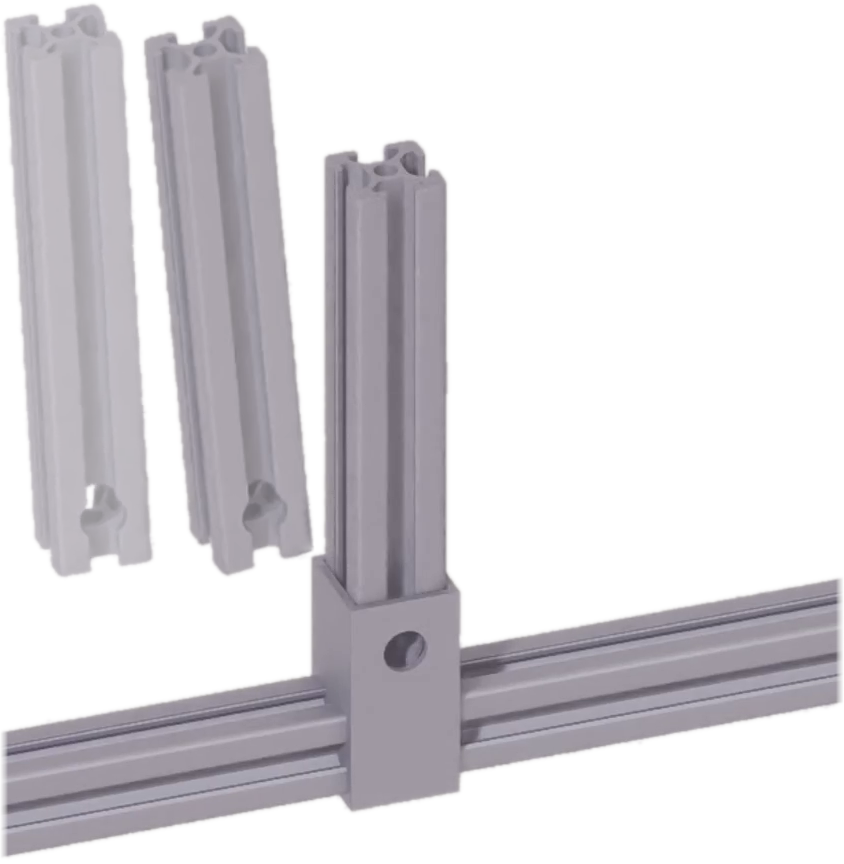}
        {\footnotesize Step 1: Profile insertion}
    \end{minipage}
    \hspace{0.5em}
    \begin{minipage}[b]{0.45\linewidth}
        \centering
        \includegraphics[width=1.0\linewidth]{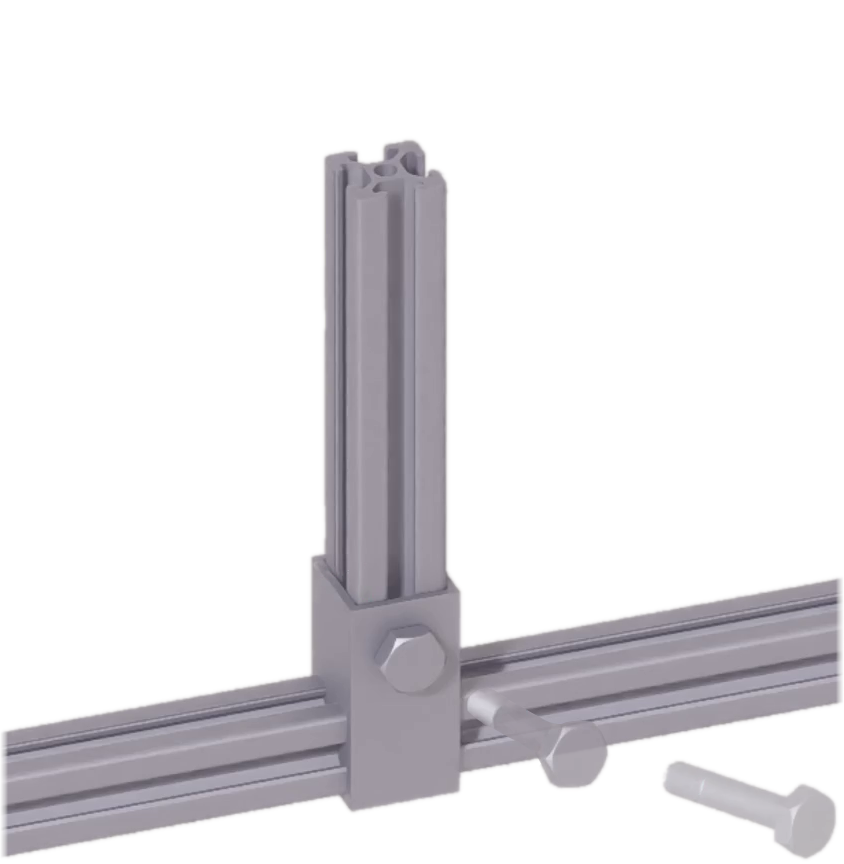}
        {\footnotesize Step 2: Bolt insertion}
    \end{minipage}
    \caption{Successful demonstration of the adaptability of our agents to a novel peg-in-hole assembly task. The task sequence consists of inserting a profile into its corresponding hole, followed by a bolt securing it in place.}
    \label{fig:assembly-task-sequence}
\end{figure}

To further evaluate the applicability of our approach, we demonstrate the generalization of the trained agents to a novel simulated scenario that reflects the challenges of real-world assembly tasks. As illustrated in Fig.~\ref{fig:assembly-task-sequence}, we consider a modular assembly sequence consisting of a profile insertion followed by a bolt insertion securing the profile in place, which is representative of the type of assembly tasks that will be encountered in space. In this demonstration, we employ the best-performing DreamerV3 agent and further challenge its adaptability by reducing the clearance between the profile and its hole to~0.25~mm, which is~25\% of the clearance that the agent encountered during training. The agent is able to successfully complete both steps of the assembly task sequence despite the increased difficulty and novel scenario. For the profile insertion part of the sequence, the agent attains a success rate of 87.64\%~(n~=~10240), with a median time until successful completion of~3.81~s. This result demonstrates the adaptability of the trained agents to novel assembly tasks and highlights the potential of our approach for addressing the challenges of space robotics.

\section{Discussion}\label{sec:discussion}

The results of our experimental evaluation provide valuable insights into the advancements and challenges of robot learning research within the context of autonomous robotic assembly. Our findings highlight the significant contribution of diverse training scenarios to the generalization capabilities of RL agents, as well as the impact of algorithmic choices on the learning process.

The evaluation of RL agents on novel peg-in-hole modules has revealed a significant impact of the training methodology on the generalization capabilities of agents. DreamerV3 achieved the highest success rate on both the training and test sets, with only a~0.38\% drop in performance when exposed to novel scenarios. This result signifies the robust generalization capabilities of DreamerV3 and the potential of model-based RL in addressing the challenges of autonomous robotic manipulation in space. On the other hand, PPO, with its stable model-free on-policy learning process, failed to systematically explore the state space and discover a successful policy, highlighting the inherent challenges of the task for this methodology.

The POMDP nature of the peg-in-hole assembly task emphasizes the critical role of temporal dependencies in the learning process. The enhanced performance of agents with observation stacking highlights the importance of incorporating temporal context in the broader field of robot learning in order to enhance the problem-solving capabilities of intelligent agents. The incorporation of temporal dependencies with even longer horizons can be facilitated through recurrent policies or even world models with a recurrent state, as demonstrated by DreamerV3. This approach could further enhance the generalization capabilities of RL agents and enable them to adapt to a wider range of novel scenarios.

Alongside the selection of a suitable RL algorithm, the diversity of training scenarios is instrumental in enhancing the generalization capabilities of RL agents. The utilization of procedurally generated peg-in-hole modules contributed to the significant performance of the trained agents in novel scenarios. In particular, the transfer of a DreamerV3 agent to a novel assembly task sequence of higher complexity resulted in a mere 6.68\% drop in success rate compared to the training set. This sets a precedent for the potential of this approach in addressing the challenges of space robotics, where agents are required to adapt to a wide range of assembly tasks despite the unpredictability of space.

\section{Conclusion and Future Directions}\label{sec:conclusion-and-future-directions}

In this work, we introduced an approach for learning autonomous peg-in-hole assembly with a focus on enhancing the generalization capabilities of RL agents. The integration of procedural generation and domain randomization in a highly parallelized simulation environment has been instrumental in training agents capable of generalizing to a wide range of novel scenarios. The comparative analysis of different RL algorithms has provided insights into their trade-offs, while the impact of temporal dependencies on the learning process has been explored. Furthermore, the adaptability of our agents was demonstrated on novel assembly sequences, showcasing the potential of the approach for space robotics.

Looking forward, the integration of procedural generation with RL holds promise for addressing the challenges of robotics in the unpredictable domain of space. Future research directions could explore the generation of procedural modules for more complex assembly tasks and sequences. Moreover, the adoption of this approach to other environments and the incorporation of visual observations could enhance the overall task performance and contribute to the development of more adaptable autonomous systems. Future efforts should also explore the impact of procedural generation and domain randomization on the transfer of trained agents to real-world assembly scenarios, with an emphasis on the safety and robustness of the learned policies.

In conclusion, this work advances the understanding of the generalization capabilities of RL agents in contact-rich manipulation tasks, with the potential to advance the assembly and maintenance of in-space infrastructure to enable more ambitious space missions and a sustainable human presence beyond Earth.



\begin{thebibliography}{10}
\providecommand{\url}[1]{#1}
\csname url@samestyle\endcsname
\providecommand{\newblock}{\relax}
\providecommand{\bibinfo}[2]{#2}
\providecommand{\BIBentrySTDinterwordspacing}{\spaceskip=0pt\relax}
\providecommand{\BIBentryALTinterwordstretchfactor}{4}
\providecommand{\BIBentryALTinterwordspacing}{\spaceskip=\fontdimen2\font plus
\BIBentryALTinterwordstretchfactor\fontdimen3\font minus
  \fontdimen4\font\relax}
\providecommand{\BIBforeignlanguage}[2]{{%
\expandafter\ifx\csname l@#1\endcsname\relax
\typeout{** WARNING: IEEEtran.bst: No hyphenation pattern has been}%
\typeout{** loaded for the language `#1'. Using the pattern for}%
\typeout{** the default language instead.}%
\else
\language=\csname l@#1\endcsname
\fi
#2}}
\providecommand{\BIBdecl}{\relax}
\BIBdecl

\bibitem{nasa2020artemis}
{National Aeronautics and Space Administration (NASA)}, ``{Artemis Plan: NASA's
  Lunar Exploration Program Overview},'' 2020.

\bibitem{zhihui2021review}
X.~Zhihui, L.~Jinguo, W.~Chenchen, and T.~Yuchuang, ``Review of in-space
  assembly technologies,'' \emph{Chinese Journal of Aeronautics}, vol.~34,
  no.~11, 2021.

\bibitem{esa2023in_orbit_servicing}
{European Space Agency (ESA)}, ``{ESA moves ahead with In-Orbit Servicing
  missions},'' 2023.

\bibitem{xu2019compare}
J.~Xu, Z.~Hou, Z.~Liu, and H.~Qiao, ``{Compare Contact Model-based Control and
  Contact Model-free Learning: A Survey of Robotic Peg-in-hole Assembly
  Strategies},'' \emph{arXiv preprint arXiv:1904.05240}, 2019.

\bibitem{elguea2023review}
{\'I}.~Elguea-Aguinaco \emph{et~al.}, ``A review on reinforcement learning for
  contact-rich robotic manipulation tasks,'' \emph{Robotics and
  Computer-Integrated Manufacturing}, vol.~81, 2023.

\bibitem{jasim2017contact}
I.~F. Jasim, P.~W. Plapper, and H.~Voos, ``{Contact-State Modelling in
  Force-Controlled Robotic Peg-in-Hole Assembly Processes of Flexible Objects
  Using Optimised Gaussian Mixtures},'' \emph{Proceedings of the Institution of
  Mechanical Engineers, Part B: Journal of Engineering Manufacture}, vol. 231,
  no.~8, 2017.

\bibitem{lee2022contact}
H.~Lee, S.~Park, K.~Jang, S.~Kim, and J.~Park, ``{Contact State Estimation for
  Peg-in-Hole Assembly Using Gaussian Mixture Model},'' \emph{IEEE Robotics and
  Automation Letters}, vol.~7, no.~2, 2022.

\bibitem{tang2016autonomous}
T.~Tang, H.-C. Lin, Y.~Zhao, W.~Chen, and M.~Tomizuka, ``{Autonomous Alignment
  of Peg and Hole by Force/Torque Measurement for Robotic Assembly},'' in
  \emph{IEEE International Conference on Automation Science and Engineering
  (CASE)}, 2016.

\bibitem{tang2016teach}
T.~Tang \emph{et~al.}, ``{Teach Industrial Robots Peg-Hole-Insertion by Human
  Demonstration},'' in \emph{IEEE International Conference on Advanced
  Intelligent Mechatronics (AIM)}, 2016.

\bibitem{wan2017optimal}
A.~Wan, J.~Xu, H.~Chen, S.~Zhang, and K.~Chen, ``{Optimal Path Planning and
  Control of Assembly Robots for Hard-Measuring Easy-Deformation Assemblies},''
  \emph{IEEE/ASME Transactions on Mechatronics}, vol.~22, no.~4, 2017.

\bibitem{cho2020learning}
N.~J. Cho, S.~H. Lee, J.~B. Kim, and I.~H. Suh, ``{Learning, Improving, and
  Generalizing Motor Skills for the Peg-in-Hole Tasks Based on Imitation
  Learning and Self-Learning},'' \emph{Applied Sciences}, vol.~10, no.~8, 2020.

\bibitem{wang2023learning}
K.~Wang, Y.~Zhao, and I.~Sakuma, ``{Learning Robotic Insertion Tasks From Human
  Demonstration},'' \emph{IEEE Robotics and Automation Letters}, 2023.

\bibitem{inoue2017deep}
T.~Inoue, G.~De~Magistris, A.~Munawar, T.~Yokoya, and R.~Tachibana, ``{Deep
  Reinforcement Learning for High Precision Assembly Tasks},'' in
  \emph{IEEE/RSJ International Conference on Intelligent Robots and Systems
  (IROS)}, 2017.

\bibitem{beltran2020variable}
C.~C. Beltran-Hernandez, D.~Petit, I.~G. Ramirez-Alpizar, and K.~Harada,
  ``{Variable Compliance Control for Robotic Peg-in-Hole Assembly: A Deep
  Reinforcement Learning Approach},'' \emph{Applied Sciences}, vol.~10, no.~19,
  2020.

\bibitem{li2022multiple}
X.~Li, J.~Xiao, W.~Zhao, H.~Liu, and G.~Wang, ``Multiple peg-in-hole compliant
  assembly based on a learning-accelerated deep deterministic policy gradient
  strategy,'' \emph{Industrial Robot}, vol.~49, no.~1, 2022.

\bibitem{wang2022deep}
F.~Wang, B.~Cui, Y.~Liu, and B.~Ren, ``{Deep Reinforcement Learning for
  Peg-in-hole Assembly Task Via Information Utilization Method},''
  \emph{Journal of Intelligent \& Robotic Systems}, vol. 106, no.~1, 2022.

\bibitem{luo2018deep}
J.~Luo, E.~Solowjow, C.~Wen, J.~A. Ojea, and A.~M. Agogino, ``{Deep
  Reinforcement Learning for Robotic Assembly of Mixed Deformable and Rigid
  Objects},'' in \emph{IEEE/RSJ International Conference on Intelligent Robots
  and Systems (IROS)}, 2018.

\bibitem{thomas2018learning}
G.~Thomas, M.~Chien, A.~Tamar, J.~A. Ojea, and P.~Abbeel, ``{Learning Robotic
  Assembly from CAD},'' in \emph{IEEE International Conference on Robotics and
  Automation (ICRA)}, 2018.

\bibitem{ding2019transferable}
J.~Ding, C.~Wang, and C.~Lu, ``{Transferable Force-Torque Dynamics Model for
  Peg-in-hole Task},'' \emph{arXiv preprint arXiv:1912.00260}, 2019.

\bibitem{hochreiter1997long}
S.~Hochreiter and J.~Schmidhuber, ``{Long Short-Term Memory},'' \emph{Neural
  Computation}, vol.~9, no.~8, 1997.

\bibitem{tobin2017domain}
J.~Tobin \emph{et~al.}, ``{Domain Randomization for Transferring Deep Neural
  Networks from Simulation to the Real World},'' in \emph{IEEE/RSJ
  International Conference on Intelligent Robots and Systems (IROS)}, 2017.

\bibitem{zhou2019continuity}
Y.~Zhou, C.~Barnes, J.~Lu, J.~Yang, and H.~Li, ``{On the Continuity of Rotation
  Representations in Neural Networks},'' in \emph{IEEE/CVF Conference on
  Computer Vision and Pattern Recognition (CVPR)}, 2019.

\bibitem{orsula2022learning}
A.~Orsula, S.~B{\o}gh, M.~Olivares-Mendez, and C.~Martinez, ``{Learning to
  Grasp on the Moon from 3D Octree Observations with Deep Reinforcement
  Learning},'' in \emph{IEEE/RSJ International Conference on Intelligent Robots
  and Systems (IROS)}, 2022.

\bibitem{blender}
\BIBentryALTinterwordspacing
{Blender Development Team}, ``{Blender 4.0}.'' [Online]. Available:
  \url{https://blender.org}
\BIBentrySTDinterwordspacing

\bibitem{omniverse}
\BIBentryALTinterwordspacing
{NVIDIA Corporation}, ``{NVIDIA Omniverse}.'' [Online]. Available:
  \url{https://nvidia.com/omniverse}
\BIBentrySTDinterwordspacing

\bibitem{macklin2020local}
M.~Macklin \emph{et~al.}, ``{Local Optimization for Robust Signed Distance
  Field Collision},'' \emph{Proceedings of the ACM on Computer Graphics and
  Interactive Techniques}, vol.~3, no.~1, 2020.

\bibitem{schulman2017proximal}
J.~Schulman, F.~Wolski, P.~Dhariwal, A.~Radford, and O.~Klimov, ``{Proximal
  Policy Optimization Algorithms},'' \emph{arXiv preprint arXiv:1707.06347},
  2017.

\bibitem{haarnoja2018soft}
T.~Haarnoja, A.~Zhou, P.~Abbeel, and S.~Levine, ``{Soft Actor-Critic:
  Off-Policy Maximum Entropy Deep Reinforcement Learning with a Stochastic
  Actor},'' in \emph{International Conference on Machine Learning (ICML)},
  2018.

\bibitem{hafner2023mastering}
D.~Hafner, J.~Pasukonis, J.~Ba, and T.~Lillicrap, ``{Mastering Diverse Domains
  through World Models},'' \emph{arXiv preprint arXiv:2301.04104}, 2023.

\bibitem{mnih2015human}
V.~Mnih \emph{et~al.}, ``\BIBforeignlanguage{en}{Human-level control through
  deep reinforcement learning},'' \emph{\BIBforeignlanguage{en}{Nature}}, vol.
  518, no. 7540, 2015.

\end{thebibliography}
\end{document}